\documentclass[10pt,twocolumn,letterpaper]{article}

\usepackage{wacv}
\usepackage{times}
\usepackage{epsfig}
\usepackage{graphicx}
\usepackage{amsmath}
\usepackage{amssymb}
\usepackage{acronym}
\usepackage{float}
\usepackage{multirow}
\usepackage[table,xcdraw]{xcolor}

\newacro{GCN}{Graph Convolutional Network}
\newacro{GC}{Graph Convolution}
\newacro{HPA}{Human Postural Assessment}
\newacro{HAE}{Human Activity Evaluation}
\newacro{HAS}{Human Activity Segmentation}
\newacro{HAR}{Human Action Recognition}
\newacro{ERA}{Ergonomics Risk Assessment}
\newacro{LSTM}{Long-Short-Term-Memory}
\newacro{TCN}{Temporal Convolutional Network}
\newacro{C3D}{3D Convolutional Neural Networks}
\newacro{REBA}{Rapid Entire Body Assessment}
\newacro{EAWS}{European Assembly Worksheet}
\newacro{ED-TCN}{Encoder-Decoder Temporal Convolutional Network}
\newacro{RNN}{Recurrent Neural Network}
\newacro{MTL}{Multi-Task Learning}
\newacro{STL}{Single-Task Learning}
\newacro{FC}{Fully Connected}

\wacvfinalcopy % *** Uncomment this line for the final submission
\ifwacvfinal
\def\assignedStartPage{1} % *** Enter the assigned starting page number (instead of 9876)
\fi

\ifwacvfinal
\usepackage[breaklinks=true,bookmarks=false]{hyperref}
\else
\usepackage[pagebackref=true,breaklinks=true,colorlinks,bookmarks=false]{hyperref}
\fi

\ifwacvfinal
\else
\fi

\begin{document}

%%%%%%%%% TITLE
\title{A Multi-Task Learning Approach for Human Activity Segmentation\\ and Ergonomics Risk Assessment}

\author{Behnoosh Parsa\hspace{2cm} 
 Ashis G. Banerjee\\
University of Washington\\
{\tt\small \{behnoosh, ashisb\}@uw.edu}}

\maketitle
%\thispagestyle{empty}

%%%%%%%%% ABSTRACT
\begin{abstract}
We propose a new approach to \ac{HAE} in long videos using graph-based multi-task modeling. Previous works in activity evaluation either directly compute a metric using a detected skeleton or use the scene information to regress the activity score. These approaches are insufficient for accurate activity assessment since they only compute an average score over a clip, and do not consider the correlation between the joints and body dynamics. Moreover, they are highly scene-dependent which makes the generalizability of these methods questionable. We propose a novel multi-task framework for \ac{HAE} that utilizes a \acl{GCN} backbone to embed the interconnections between human joints in the features. In this framework, we solve the \ac{HAS} problem as an auxiliary task to improve activity assessment. The \ac{HAS} head is powered by an \acl{ED-TCN} to semantically segment long videos into distinct activity classes, whereas, \ac{HAE} uses a \acl{LSTM}-based architecture. We evaluate our method on the UW-IOM and TUM Kitchen datasets and discuss the success and failure cases in these two datasets.
\end{abstract}
\vspace{-.5em}
%%%%%%%%% BODY TEXT
%////////////////////////////////////////////////////////////////////////
\section{Introduction}
\noindent
With the advancements in computer vision techniques, automated \acf{HAE} has received significant attention. The aim of this category of problems is to design a computational model that captures the dynamic changes in human movement and measures the quality of human actions based on a predefined metric.
\ac{HAE} has been studied in a variety of computer vision applications such as sports activity scoring, athletes training \cite{patrona2018motion, weeratunga2017application, parmar2019and}, rehabilitation and healthcare \cite{parmar2016measuring,baptista2017video}, interactive games \cite{zhang2017martial,meng2018distances}, skill assessment \cite{li2019manipulation, doughty2018s}, and workers activity assessment in industrial settings \cite{parsa2019toward, parsa2020spatio}. 
\begin{figure}
\centering
  \includegraphics[width=0.5\textwidth]{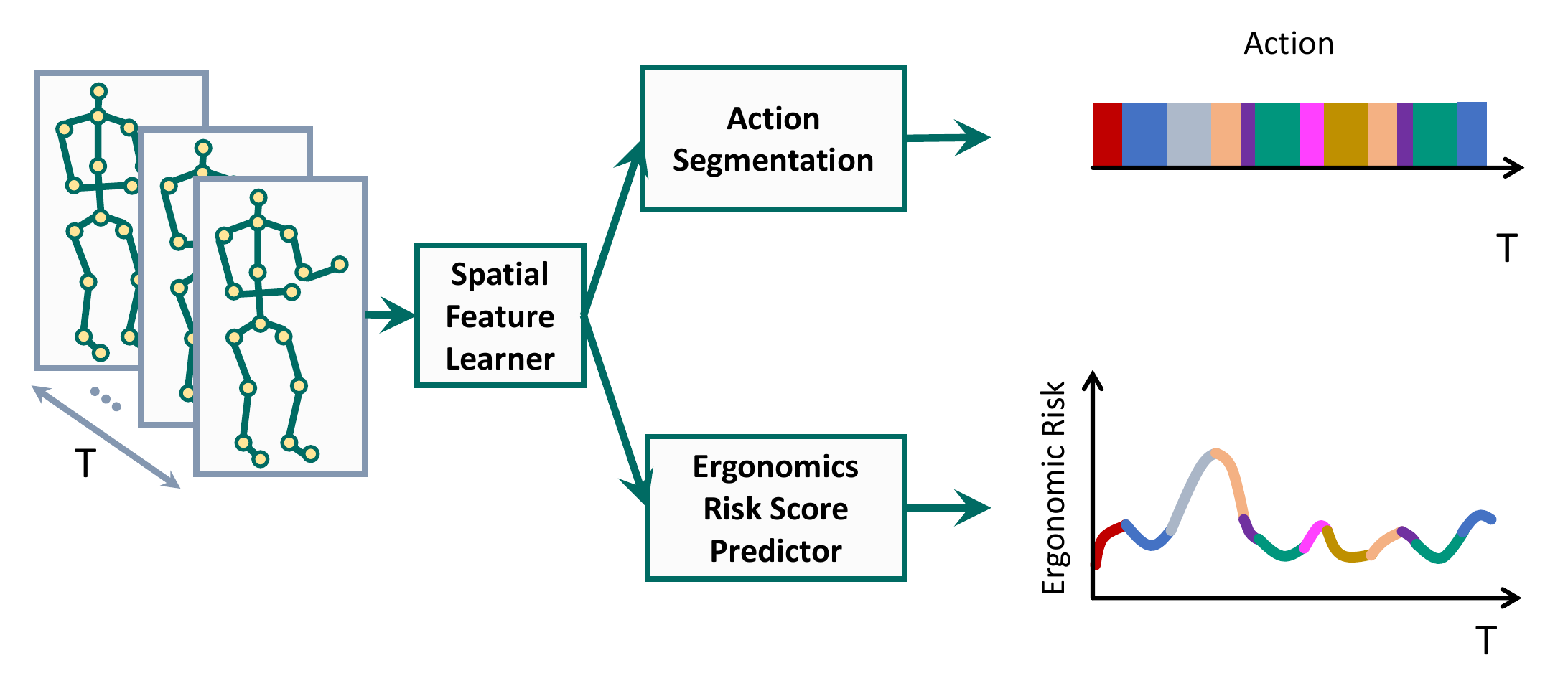}%
  \caption{Multi-task activity segmentation and ergonomics risk assessment pipeline.}
  \label{fig:pipeline}
\end{figure}
Some of the earlier works on \ac{HAE} used traditional feature extraction methods for performance analysis \cite{pirsiavash2014assessing, ilg2003estimation}. Recently, with the popularity of deep learning methods, a multitude of creative solutions have emerged for solving \ac{HAE} problems. Among the proposed methods, some directly learn a mapping from images to a quality score \cite{xiang2018s3d}.  As the activity quality is highly task-dependent a majority of research is focused on leveraging the available activity information in the learning process \cite{parmar2019and, parsa2020spatio}. Another approach has been to measure the deviation of a test sequence from a template sequence for determining the activity quality \cite{parisi2016human}. This approach is valuable when the performance of humans is evaluated based on how well they followed a fixed series of activities in a certain way such as in sport competitions or manufacturing operations.

There is another aspect of \ac{HAE} that has received less attention despite its importance and potential impact on the safety and health of the society. \acf{HPA} is studied in various fields such as biomechanics, physiotherapy, neuroscience, and more recently in computer vision \cite{parsa2019toward, parsa2020spatio, malaise2019activity}. \ac{HPA} is a subcategory of \ac{HAE} that focuses on determining the quality of human posture using a ergonomics-based (or biomechanics-based) criteria. There are three major challenges in solving \ac{HPA} problems: (1) the type of task and the object involved in the activity highly influence the risk level. (2) The repetition of certain movements can cause accumulated pressure on specific body parts. Therefore, it is important to analyze a video in a frame-wise fashion to be able to capture repetition. (3) Everyone does not necessarily perform a task in the same way, hence, a successful algorithm should learn the relation between human joints dynamics and the corresponding ergonomics risk score.

This work is inspired by the importance of \ac{HPA} problems and their significant impact on the health and safety of industrial workers. However, our approach is not limited to this specific application and it is a novel design that can benefit other aspects of \ac{HAE} research. We leverage from consistent representation of human 3D pose and propose an end-to-end multi-task framework (Figure \ref{fig:pipeline}) that solves \acf{HAS} as an auxiliary task to improve the \ac{HPA} performance. Skeleton-based methods have been shown to provide the opportunity of developing more generalizable algorithms for various applications in \ac{HAR} and prediction problems~\cite{rogez2017lcr}. However, they have not been leveraged enough in \ac{HAE}.
% The feature extraction backbone for both branches is based on a Graph Neural Network design to capture the inter relation between human joints. 
%-------------------------------------------------------------------------

%\noindent
\textbf{Contributions:} This work brings together activity segmentation and activity assessment using a novel multi-task learning framework. Our proposed framework comprises a \acf{GCN} backbone and an \acf{ED-TCN} for the activity segmentation head and a \acf{LSTM}-based head for activity assessment. The contribution of our work is threefold. (1) We introduce a novel combination of \ac{GCN} with \ac{ED-TCN} for activity segmentation in long videos that outperforms state-of-the-art results on the UW-IOM dataset. (2) Our \ac{MTL}-emb method initiates a line of research for more informed activity assessment by fusing activity embedding with spatial features for \ac{ERA}. (3) We present a way to use the skeletal information for activity assessment in a \acf{MTL} framework that may enable generalization across a variety of environments and leverage anthropometric information. 
% To the best of our knowledge, this is the first attempt at using skeletal data in a \ac{MTL} for \ac{HAE}.
%
% More specifically, we answer the following questions in this paper:
% \begin{itemize}
% \item How well does end-to-end activity segmentation perform compared to the methods that use features from pre-trained networks?
% \item Is the 15 joint position sufficient representation for action segmentation task, and can a \ac{GCN} compete with context based features like VGG16 to provide spatial features for the activity segmentation task? 
% \item Does a multi task approach enhance ergonomics risk assessment?
% \item How does the fusion of action embedding affect multi-task performance? 
% \end{itemize}
%-------------------------------------------------------------------------
%////////////////////////////////////////////////////////////////////////
\section{Related Work}
\noindent
\ac{HAS} is the task of semantically segmenting a video into clips corresponding various activities and localizing their start and end times. \ac{HPA} considers the task of finding the ergonomics risk score corresponding to the human posture at every frame of a video. To the best of our knowledge, this is the first work that combines the two separately studied problems of \ac{HAS} and \ac{HPA} in a multi-task setting. Moreover, the combination of the \ac{GCN} backbone with a powerful \ac{ED-TCN} structure for \acl{STL}-based \ac{HAS} (STL-AD) is a novel idea that can compete with methods using image-based features (if the actions are not too similar). The \ac{HPA} branch also offers a new combination of \ac{GCN} along with a \ac{LSTM} unit to learn the relation between human joint dynamics and the corresponding ergonomics risk score. In this section, we summarize the works related to \ac{HAE}, \ac{GCN}, and \ac{ERA} methods to provide the background for our proposed solution.
%-------------------------------------------------------------------------
% \vspace{-.5em}
\subsection{Human Activity Evaluation}
\noindent
Also known as Action Quality Assessment, \ac{HAE} focuses on designing models that are able to learn a mapping between human body dynamics and the completion quality of the performed actions based on an accepted metric or a template sequence (refer to \cite{lei2019survey} for further literature on early methods with handcrafted features). The majority of deep learning approaches to HAE have focused on using \ac{C3D} \cite{tran2015learning} and Pseudo-3D Networks (P3D) \cite{xiang2018s3d} to extract spatio-temporal features that are fed into a regression unit. One of the recent works in applications for physical therapy, \cite{liao2020deep}, proposed a framework including performance metrics, scoring functions, and different neural network architectures for mapping joint coordinates to the activity score. Similarly, \cite{parmar2017learning} used C3D to extract spatio-temporal features and conducted performance score regression using a \ac{LSTM} unit for data from Olympic events. Despite the value of all these works in initiating the use of computer vision techniques for \ac{HAE} in rehabilitation and sports, the proposed methods are highly dependent on the context of the video frames. 
%and thus lack generalizability. 
Moreover, the learned mapping between the frames and the score does not incorporate the effect of human body kinematics.

Recently, there have been efforts in leveraging human body kinematics in designing deep architectures for evaluating surgical skills \cite{fawaz2018evaluating}. This work uses 75 dimensional kinematic data (3D coordinates plus velocities) of two surgical tools being manipulated by surgeons and classifies the skill level into expert, intermediate, and novice.  Joint relation graph has been utilized to assess the performance of athletes in Olympic events \cite{pan2019action}. The proposed joint relation graph is a spatial \ac{GCN} with node features that are outputs of I3D \cite{carreira2017quo} on image patches containing the human joints.

 Parmar and Morris's work \cite{parmar2019and} is the most similar work to our paper. They propose a multi-task framework utilizing spatio-temporal features to solve action recognition, commentary generation, and \ac{HAE} score estimation for Olympic events. However, the focus of their work is on short video classification, where each clip includes only one activity, namely, diving of one athlete. In contrasts, our focus is on localizing actions in a long video while simultaneously inferring the ergonomics risk of human posture at every frame. 
% Moreover, by bringing our novel attention-based \ac{GCN} into a multitask-HAE framework we are able to sheds light on the joints that play important role in the resultant risk score.
%------------------------------------------------------------------------
\subsection{Graph Convolution Networks}
\noindent
\ac{GCN}s was developed to process data belonging to non-Euclidean spaces \cite{wu2019comprehensive}. \ac{GCN}s are the most intuitive choice for human body kinematics since the commonly-used independent and identically distributed random variable assumption is not applicable. Spatio-Temporal Graph Convolutional Networks (ST-GCN) introduced a powerful tool for analyzing human motions in videos, and has been utilized in several computer vision applications \cite{yan2018spatial,kim2019skeleton,li2019actional,si2019attention}. However, most of these works focus on solving \acf{HAR} problems.
Recently, \cite{parsa2020spatio} introduced a Spatio-Temporal Pyramid Graph Network (ST-PGN) for early action recognition. They also used the predicted activity labels to enhance \ac{ERA} that was computed using 3D skeletal reconstruction. In this work, we leverage a GCN backbone to learn the joint embedding and use that to directly predict the ergonomics risks rather than solving it as a separate problem. %independent of the action recognition module.
%------------------------------------------------------------------------
\subsection{Ergonomics Risk Assessment}
\noindent
The United States alone has annually more than 150,000 workers suffering from back injuries due to repetitive lifting of heavy objects with inappropriate postures. Hence, many studies have recently looked at designing automated \ac{ERA} methods \cite{peppoloni2016novel, singh2017ergonomic, colim2019ergonomic, roodbandi2018prevalence, possebom2018comparison, malaise2019activity, parsa2019toward, parsa2020spatio}. The most widely used methods in the industry are \ac{REBA} \cite{hignett2004rapid} and \ac{EAWS} \cite{schaub2013european}. \ac{REBA} provides a risk score between 1-15 by considering all the main body joint angles, magnitude of the applied force, and ease of grasping an object. \ac{EAWS} is a similar method that focuses on the upper extremity postures in assembly tasks. In practice, the quantification of risk values is mostly based on observations.

Automated \ac{ERA} research can be broadly divided into two main categories. One line of research focuses on reducing ergonomics risk in a collaborative setting, where a robot has to place the work platform in a configuration that minimizes the ergonomics risk \cite{maurice2019human, Shafti2019}. Others have used body mounted sensors to measure kinematics for real-time \ac{ERA} \cite{li2011computer,malaise2019activity}.
Another line of research focuses on learning ergonomics risk for various individual actions. In \cite{parsa2019toward}, the \ac{ERA} problem is taken as an action localization problem and \ac{TCN} is used to segment the videos into tasks with different risk labels. The ergonomics risk is computed offline and the dataset is labeled with high-, medium-, and low-risk labels. In addition, a dataset on common industry-related activities is introduced in \cite{parsa2019toward} that we use to evaluate the performance of our proposed method. In \cite{parsa2020spatio} the problem is approached as an action recognition problem on long videos, and the predicted activity class is used to modify the computed ergonomics risk through a parallel algorithm. This work, on the other hand, introduces a multi-task \ac{HPA} framework that predicts ergonomics risk directly from human pose with the help of \ac{HAS} as an auxiliary task. %TODO
%////////////////////////////////////////////////////////////////////////
\begin{figure}[t]
\centering
% \fbox{\rule{0pt}{2in} \rule{.9\linewidth}{0pt}}
\includegraphics[width=0.8\linewidth]{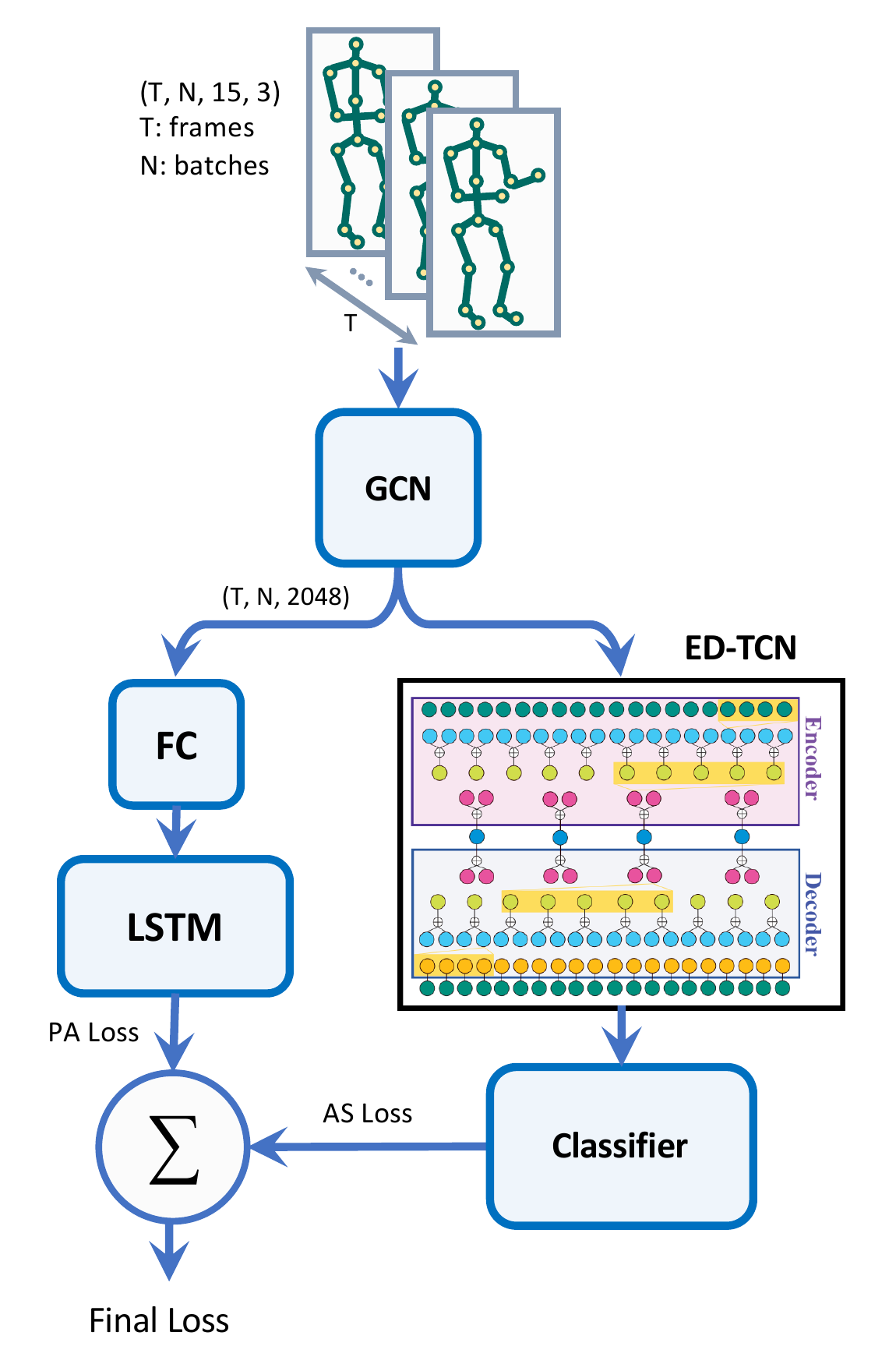}
\caption{MTL network architecture.}
\label{fig:D-pipeline}
\end{figure}
\section{Proposed Multi-Task Framework}
\noindent
In \ac{ERA}, posture alone cannot accurately determine the risk level. The activity class contains information that is key to measure ergonomics risk. We, therefore, define \ac{HPA} as a \ac{MTL} problem consisting of an \ac{HAS} and an \ac{HPA} task (Figure \ref{fig:D-pipeline}). In the following sections, each component of our \ac{MTL} model is described in details.
%-------------------------------------------------------------------------
\subsection{Spatial Features}
\noindent
The inputs to our multi-task model are 3D joints locations, which is a form of structured data. Since \ac{GCN}s are known to be powerful in representing structured data \cite{zhou2018graph}, our model uses a sequence of stacked \ac{GCN}s as the backbone for spatial feature extraction, similar to the proposed structure in \cite{yan2018spatial} except for temporal convolution. Just like a 2D convolutional layer, a stacked \ac{GCN} allows better feature extraction for unstructured data such as graphs.

Given the input $\mathbf{x} \in \mathbf{R}^{D\times N}$, where $D$ is equal to 3 as the joints are represented using $(x, y, z)$ coordinates and $N$ is the number of joints, the adjacency matrix $\mathbf{A} \in \mathbf{R}^{N \times N}$, and the degree matrix $\hat{\mathbf{D}}$ with
$\mathbf{D}_{i i}=\sum_{j} \mathbf{A}_{i j}$, a \ac{GC} can be written as,
%%%%%%%%%%%%%%%%%%%%%%%%%%%%%%%%% Eq %%%%%%%%%%%%%%%%%%%%%%%%%%%%%%%%%%%%%%
\begin{equation}\label{eq:graph1}
    \mathbf{f}=\hat{\mathbf{D}}^{-\frac{1}{2}} \hat{\mathbf{A}} \hat{\mathbf{D}}^{-\frac{1}{2}} {\mathbf{x}}^{\top} \mathbf{W} \ .
\end{equation}
%%%%%%%%%%%%%%%%%%%%%%%%%%%%%%%%%%%%%%%%%%%%%%%%%%%%%%%%%%%%%%%%%%%%%%%%%%%
Here, $\hat{\mathbf{A}}=\mathbf{A}+\mathbf{I}$, $\mathbf{I}$ is the identity matrix. For a graph with human skeletal structure, $\mathbf{A}$ is designed based on the anatomical connections among the joints. $\mathbf{W} \in \mathbf{R}^{D \times F}$ is the weight matrix that is to be learned. Hence, if the input to a \ac{GCN} layer is $D\times N$, the output feature $\mathbf{f}$ is $N\times F$, where $F$ is the chosen output feature size. In our proposed backbone, each \ac{GCN} is followed by a ReLU activation.
Moreover, the adjacency matrix is partitioned into three sub-matrices as described in \cite{yan2018spatial} to better capture the spatial relations among the joints. Therefore, Equation~(\ref{eq:graph1}) is written in a summation form for each \ac{GCN} layer as:
%%%%%%%%%%%%%%%%%%%%%%%%%%%%%%%%% Eq %%%%%%%%%%%%%%%%%%%%%%%%%%%%%%%%%%%%%%
\begin{equation}\label{eq:graph2}
    \mathbf{f}=\sum_{a=1}^3\hat{\mathbf{D}}_a^{-\frac{1}{2}}\mathbf{A}_a \hat{\mathbf{D}}_a^{-\frac{1}{2}} {\mathbf{x}}^{\top} \mathbf{W}_a \ ,
\end{equation}
%%%%%%%%%%%%%%%%%%%%%%%%%%%%%%%%%%%%%%%%%%%%%%%%%%%%%%%%%%%%%%%%%%%%%%%%%%%
where $a$ indexes each partition.
%-----------------------------------------------------------------------%
%
\begin{figure*}[t]
\centering
% \fbox{\rule{0pt}{2in} \rule{.9\linewidth}{0pt}}
\includegraphics[width=0.9\linewidth]{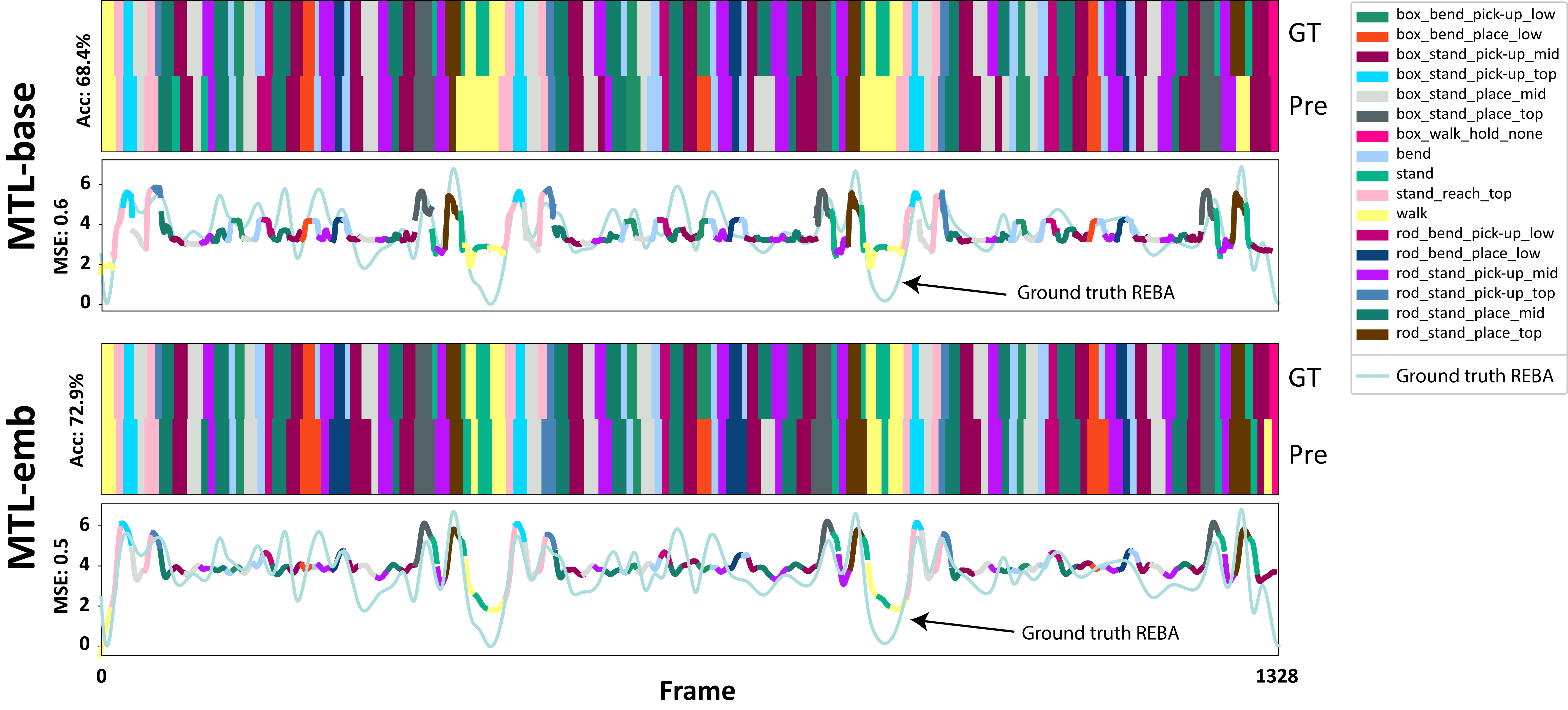}
\caption{Visualization of \ac{HAS} and REBA prediction result for a sample test video of UW-IOM dataset. The first and third plots (colored ribbons) are the segmentation results. In each ribbon the top-half is the ground truth and the bottom-half is the predictions by the network. The second and fourth plots depicting the ground truth REBA score and the network prediction. The network prediction is color-coded based on the activity class.}
\label{fig:results}
\end{figure*}
%-----------------------------------------------------------------------%
%------------------------------------------------------------------------
\subsection{Encoder-Decoder Temporal Convolution for \acl{HAS}}
\noindent
In \ac{HAS} problems, the task is to identify the activities that are happening in untrimmed videos and determine the corresponding initial and final frames \cite{gammulle2020fine, lea2017temporal, bai2018empirical, parsa2019toward}. A popular approach that is inspired by works in audio generation and speech recognition \cite{oord2016wavenet,xiong2018microsoft} is to use feed-forward (i.e., non-recurrent) networks for modeling long sequences. The main component of these methods is a 1D dilated causal convolution that can model long-term dependencies. 

A dilated convolution is a filter that applies to an area larger than its length by skipping input values by a certain length \cite{oord2016wavenet}. A causal convolution is a 1D convolution which ensures the model does not violate the ordering of the input sequence. The prediction emitted by a causal convolution (that is $p(x_{t}|x_1, ..., x_{t-1})$) at time step $t$ only depends on the previous data. Combining these two properties, dilated causal convolutions have large receptive fields and are faster than \acp{RNN}. Moreover, they are shallower than regular causal convolution due to dilation.

For the \ac{HAS} task, inspired by \cite{oord2016wavenet, lea2017temporal, parsa2019toward} we design an \ac{ED-TCN}-based on 1D dilated convolutions (Figure  \ref{fig:D-pipeline}). Our design consists of a hierarchy of four temporal convolutions, pooling, and upsampling layers. The output of the \ac{ED-TCN} followed by a \ac{FC} layer and a \texttt{RelU} activation is fed to the classification layer.

In using \ac{ED-TCN} for \ac{HAS} \cite{lea2017temporal, parsa2019toward}, the focus is on learning the temporal sequence and localizing activities. It is common to extract spatial features prior to training from an independent network like VGG16 \cite{simonyan2014very} or ResNet \cite{he2016deep}. Our proposed framework learns the spatial and temporal properties of the data in an end-to-end fashion. To our knowledge, this is the first attempt to use \ac{ED-TCN} in an end-to-end architecture with a spatial feature detector. In addition, the combination of \ac{GCN} with \ac{ED-TCN} for solving \ac{HAS} is a novel approach and it shows promising results.
%------------------------------------------------------------------------
\subsection{Regression Module for \acl{HPA}}
\noindent
We define \ac{HPA} as a sub-category of \ac{HAE} where the activity score is determined based on the safety of the posture.
In \ac{HPA}, the task is to find a mapping between the spatio-temporal features and ergonomics risk score. Our proposed regressor uses the shared spatial features coming from the \ac{GCN} backbone. The \ac{GCN} features go through a \ac{FC} layer with \texttt{tanh} nonlinearity and are then fed into a stacked \ac{LSTM} structure to predict the \ac{REBA} scores.
%------------------------------------------------------------------------
\subsection{Multi-Task Approach to \ac{ERA}}
\noindent
\ac{MTL} is a popular framework for end-to-end training of a single network for solving multiple related tasks. In these networks, a common backbone provides the data representation for branches responsible for learning a specific task. Usually in \ac{MTL}, there is a main task plus multiple auxiliary tasks that complement the core task. For instance, in \ac{HAE}, the main task is to determine the action quality. However, action quality is not independent of what action is carried out, which makes the \ac{HAS} choice of auxiliary tasks natural for this kind of problems.

The supervision signals from the auxiliary tasks can be viewed as inductive biases \cite{caruana1997multitask} that limit the hypothesis search space and result in a more generalizable solution. The multi-task approach to \ac{HAE} has been recently introduced by \cite{parmar2019and} for determining the quality of action in short clips from Olympic games.

In our work, the main task is to predict the \ac{REBA} scores. However, the information about human action is closely related to its corresponding ergonomics risk. Therefore, the auxiliary task in this case is the \ac{HAS}. The long duration videos pose an additional challenge, since, unlike most of the \ac{HAE} datasets, both the activities and their risk scores vary over time. In a majority of sport \ac{HAE} \cite{parmar2019and}, a single activity score is predicted for a clip. Here, the \ac{HAS} task consists of 17 and 20 actions for the UW-IOM and TUM datasets, respectively (see Section 4 for more information on the datasets). Therefore, in any video, activity localization and \ac{ERA} task involves predicting a smooth function that shows how the risk is changing throughout the video. 

We studied two different architectures for solving this \ac{MTL} problem. In the first architecture, the heads corresponding to each task only share the \ac{GCN}-driven features. In the second architecture, the output of the \textit{Softmax} layer of the \ac{HAS} head is fused to the feature going to the \ac{LSTM} regressor.

We consider a weighted average of the \ac{HAS} loss and the \ac{HPA} loss as the overall multi-task \ac{HPA} loss function, 
%%%%%%%%%%%%%%%%%%%%%%%%%%%%%%%%% Eq %%%%%%%%%%%%%%%%%%%%%%%%%%%%%%%%%%%%%%
\begin{equation}\label{eq:ERA_loss}
    \mathcal{L}_{HPA}= \sum_{t=1}^T{\alpha (\mathbf{x}_t - \mathbf{y}_t)^2 + \beta \vert \mathbf{x}_t - \mathbf{y}_t \vert},
\end{equation}
%%%%%%%%%%%%%%%%%%%%%%%%%%%%%%%%%%%%%%%%%%%%%%%%%%%%%%%%%%%%%%%%%%%%%%%%%%%
where $\mathbf{y}_t$ is the frame-wise ground truth REBA score and $\mathbf{x}_t$ is the model prediction. $\vert\cdot\vert$ is the $\mathcal{L}_1$ norm. $\alpha$ and $\beta$ are weights to be learned. For \ac{HAS}, we use cross-entropy loss between ground truth and model prediction,
%%%%%%%%%%%%%%%%%%%%%%%%%%%%%%%%% Eq %%%%%%%%%%%%%%%%%%%%%%%%%%%%%%%%%%%%%%
\begin{equation}\label{eq:Det_loss}
    \mathcal{L}_{HAS}= -\sum_{t=1}^T{\sum_{c=1}^{Cl}{\mathbf{y}_{t,c} \log(\mathbf{x}_{t,c})}},
\end{equation}
%%%%%%%%%%%%%%%%%%%%%%%%%%%%%%%%%%%%%%%%%%%%%%%%%%%%%%%%%%%%%%%%%%%%%%%%%%%
where $Cl$ is the number of classes. The overall loss is the sum of all the losses, 
%%%%%%%%%%%%%%%%%%%%%%%%%%%%%%%%% Eq %%%%%%%%%%%%%%%%%%%%%%%%%%%%%%%%%%%%%%
\begin{equation}\label{eq:Det_loss}
    \mathcal{L}_{MTL}= \mathcal{L}_{HPA} + \gamma \mathcal{L}_{HAS},
\end{equation}
%%%%%%%%%%%%%%%%%%%%%%%%%%%%%%%%%%%%%%%%%%%%%%%%%%%%%%%%%%%%%%%%%%%%%%%%%%%
where $\gamma$ is to be learned.
%////////////////////////////////////////////////////////////////////////
\section{Experiments}
%------------------------------------------------------------------------
\subsection{Datasets}
\noindent
Despite the impact of automated \ac{ERA} on industry, research in this area has started gaining popularity only recently. As a result, only a few datasets are available that capture representative activities in industrial settings. In particular, two such datasets have been used in recent publications in this domain.

\textbf{UW-IOM Dataset} is a publicly available dataset of 20 videos by \cite{parsa2019toward} that captures industry-relevant activities. This dataset has 17 action classes and labels are of four-tier hierarchy indicating the object, human motion, type of object manipulation (if applicable), and the relative height of the surface on which the activity is taking place. The longest video in this dataset has $2,384$ frames. We use the 3D poses for UW-IOM dataset from our earlier work \cite{parsa2020spatio}. 

\textbf{TUM Kitchen Dataset} has 19 videos consisting of daily activities in a kitchen. Learning with graph-based methods has been shown to be challenging on this datasets due to the similarity of human postures in multiple action classes \cite{parsa2020spatio}. We took labels provided by \cite{parsa2019toward} so that we can compare our results with theirs. We used \cite{pavllo:videopose3d:2019} to extract the 3D poses from the videos recorded by the second camera. The longest video in this dataset has $2,048$ frames. 

The input features to our model are 3-dimensional key-points $(x,y,z)$ of $N=$\textit{15} joints, concatenated over time $T$. Hence, the resulting input tensor is of  dimension $3 \times 15 \times T$. The output ground truth labels are frame-wise labels that have the dimension of $1 \times T$.
%------------------------------------------------------------------------
\vspace{-.4em}
\subsection{Ergonomics Risk Pre-processing}
\noindent
\ac{REBA} method \cite{hignett2004rapid} computes a score describing the total body risk based on the joint angles and the properties of an action. The \ac{REBA} scores are discrete integers from 1 (the minimum risk level) to 15 (the maximum risk level).
In \cite{parsa2019toward}, the scores of all the subjects are averaged over the classes and a single score is reported for each activity class. We used the detected skeletons to compute the joint angles and obtained a frame-wise \ac{REBA} score. However, the REBA profile then becomes a sequence of piece-wise constants, which is hard to learn by a regressor. Therefore, we smoothed the \ac{REBA} sequence using the Python \texttt{UnivariateSpline} function to make it easier for the \ac{ERA} regressor to learn the patterns. To help advance research in this area, the smoothed REBA scores along with the code are available on the project repository\footnote{https://github.com/BehnooshParsa/MTL-ERA}.
% \footnote{https://github.com/BehnooshParsa/MTL-ERA}.
%-----------------------------------------------------------------------%
\begin{figure}[t]
\centering
% \fbox{\rule{0pt}{2in} \rule{.9\linewidth}{0pt}}
\includegraphics[width=.9\linewidth]{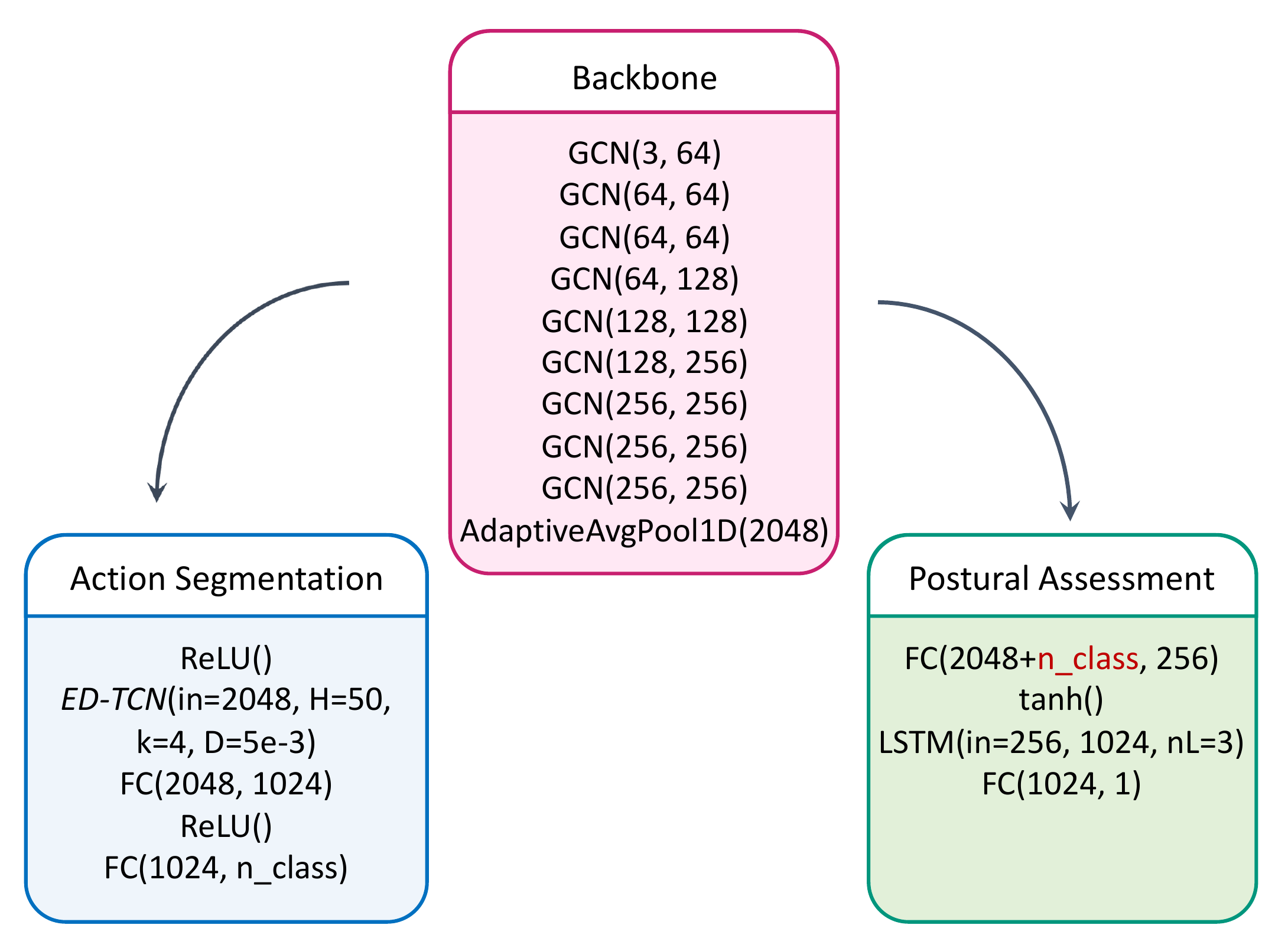}
\caption{Detailed MTL-emb architecture. $GCN(in, out)$ is a \ac{GCN} with edge-importance. ED-TCN has 4 hidden layers of size $H$ with kernel size $k$ and dropout of $D$. $FC(in, out)$ is a fully connected layer. $n_{class}$ is the number of classes. The LSTM has $nl$ layers.}
\label{fig:detailofNet}
\end{figure}
\subsection{Implementation Details}
\noindent
All the networks were implemented in PyTorch \cite{paszke2017automatic}. The initial values of the loss function weight parameters $\alpha$, $\beta$, and $\gamma$ were set to 1. All the networks were trained using the Adam optimizer \cite{kingma2014adam}. We implemented early-stopping and trained the model with different learning rates to find the best one (the best performing learning rate is shown in Table \ref{tab:MTL}). The 20 videos in the UW-IOM dataset were randomly split into 15 and 5 for the training and validation set, respectively. For the TUM dataset, the training and validation sets include 15 and 4 videos, respectively.

\textbf{GCN Backbone:} The details of the \ac{GCN} network is displayed in Figure \ref{fig:detailofNet}. The output of the final \ac{GCN} layer is of size $(N, T, 256, 15)$ that is flattened to $(N, T, 3840)$ and passed through an adaptive pool layer. Therefore, the feature that is fed to the rest of the network is of size $(N, T, 2048)$. 

\textbf{Action Segmentation Head:} \ac{ED-TCN} requires input batches to have the same temporal length. Hence, we defined a maximum length in both the training and validation sets, and masked the rest of the inputs with a value of $-1$ (thus, $T$ corresponds to the maximum sequence length). The predicted sequence was unmasked before calculating the loss. The ED-TCN output goes through two fully connected layers with \texttt{RelU} activation and is used to compute the cross-entropy loss. 

\textbf{Postural Assessment Head:} We evaluated the performance of two architectures for \ac{HPA}. In one design, we fuse the Softmax output of the \ac{HAS} head to the \ac{GCN} features and call this model, \textit{multi-task-emb}. The base design does not include fusion and we refer to that as \textit{multi-task-base}. The spatial features (from the \ac{GCN} backbone) are followed by a fully connected layer with \texttt{tanh} activation and sent to three layers of \ac{LSTM}. The LSTM output is followed by a fully connected layer to predict the REBA scores and is sent to the regression loss function.
\subsection{Evaluation Metrics}
To measure the performance of the \ac{HAS}  network we use \textit{F1-overlap score}, \textit{segmental edit score}, and \textit{Mean Average Precision} (MAP). F1-overlap score is essentially the harmonic mean of $\mathit{Precision}$  and $\mathit{Recall}$  and is computed using the following well known formula:
%%%%%%%%%%%%%%%%%%%%%%%%%%%%%%%%% Eq %%%%%%%%%%%%%%%%%%%%%%%%%%%%%%%%%%%%%%
\begin{equation}\label{eq:F1}
    \mathit{F_1\mbox{-}Score}= 2\times \frac{\mathit{Precision} \times \mathit{Recall}}{\mathit{Precision} + \mathit{Recall}} .
\end{equation}
%%%%%%%%%%%%%%%%%%%%%%%%%%%%%%%%%%%%%%%%%%%%%%%%%%%%%%%%%%%%%%%%%%%%%%%%%%%
Edit score measures the closeness of the predicted sequence to the ground truth sequence. This metric penalizes if the order of the sequence and the number of action segments are not correct. The average precision is computed over all the classes and its mean is reported.
%////////////////////////////////////////////////////////////////////////
\section{Results and Discussion}
To evaluate the strength of our proposed multi-task approach in solving the \ac{HAS} and \ac{HPA} problems, we carry out two single-task experiments for the \ac{HPA} task (STL-PA) and the \ac{HAS} task (STL-AS). 
Another reason behind the STL-AS experiment is to investigate the power of our \ac{GCN} model as a spatial feature extractor in solving \ac{HAS} problems.
% \subsection{Single-task Postural Assessment}

The STL-PA network has identical GCN backbone and LSTM design as the MTL network. The average MSE result is reported for the validation set in Table \ref{tab:STL-Reg}. It is clear from the results that the network cannot learn the sophisticated pattern of the REBA profile.
%-----------------------------------------------------------------------%
\begin{table}[H]
\centering
\includegraphics[width=.85\columnwidth]{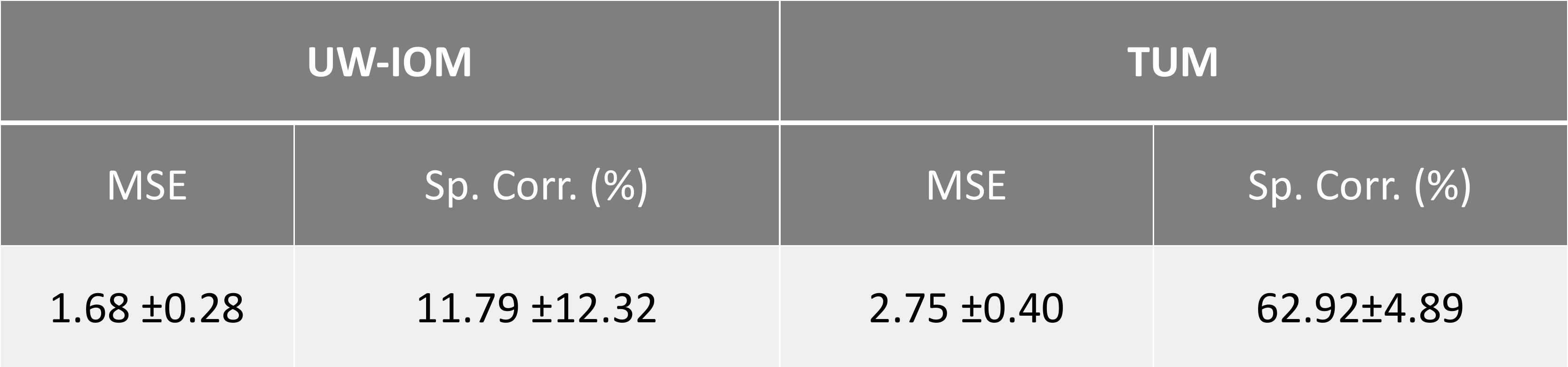}
\caption{Average MSE and Spearman's Coefficient of the activity score prediction over the validation videos using the STL-PA model.}
\label{tab:STL-Reg}
\end{table}
%-----------------------------------------------------------------------%
\subsection{Action Segmentation with GCN-ED-TCN}
\noindent
As discussed in Section 2, \ac{ED-TCN} along with the input features derived from pre-trained networks, have been widely used for \ac{HAS}. The idea is that given the input spatial features for every time-step of a sequence, this method can segment it into semantically similar pieces. Nonetheless, an end-to-end approach for learning both the spatial and temporal features in an \ac{HAS} has not been explored with \ac{ED-TCN}. While \ac{GCN} models have been used both for activity classification \cite{yan2018spatial,kim2019skeleton} and early action recognition \cite{parsa2020spatio}, its capability has not been evaluated for \ac{HAS}.
ED-TCN is used for \ac{HAS} on the UW-IOM dataset in \cite{parsa2019toward}, where the authors compare three spatial feature extractors, namely, a pre-trained VGG16 on ImageNet \cite{deng2009imagenet}, a fine-tuned version of VGG16 model, and a P-CNN model \cite{cheron2015p}. Our proposed \ac{GCN} backbone extracts spatial features based on human pose only, but its performance is comparable with the state-of-the-art as shown in Table \ref{tab:STL-Class}. Hence, we believe that pose-based features are more suitable for designing a generalizable algorithm. However, we should emphasize that generalizability comes with a price of the model not performing well when the pose information is poor or when the activities require similar postures, which is the case for the TUM dataset (Table \ref{tab:STL-Class}).
\subsection{Single-task vs. Multi-task Approach}
\noindent
%, which is substantially higher than both the MTL designs as shown in Table \ref{tab:MTL}.
%
The substantial improvement in predicting the activity risk scores is evident when comparing the results in Table \ref{tab:STL-Reg} with Table \ref{tab:MTL}. We believe that the underlying reason behind this observation is that the REBA score is highly dependent on the type of activity, and learning an auxiliary \ac{HAS} task can enhance the performance of the \ac{HPA} head. However, the inverse dependency is not that strong. Our findings indicate that the STL-AS performs better than the MTL approach for \ac{HAS} (comparing results in Table \ref{tab:STL-Class} and \ref{tab:MTL}).
%-----------------------------------------------------------------------%
\begin{table*}[]
\centering
\includegraphics[width=.85\textwidth]{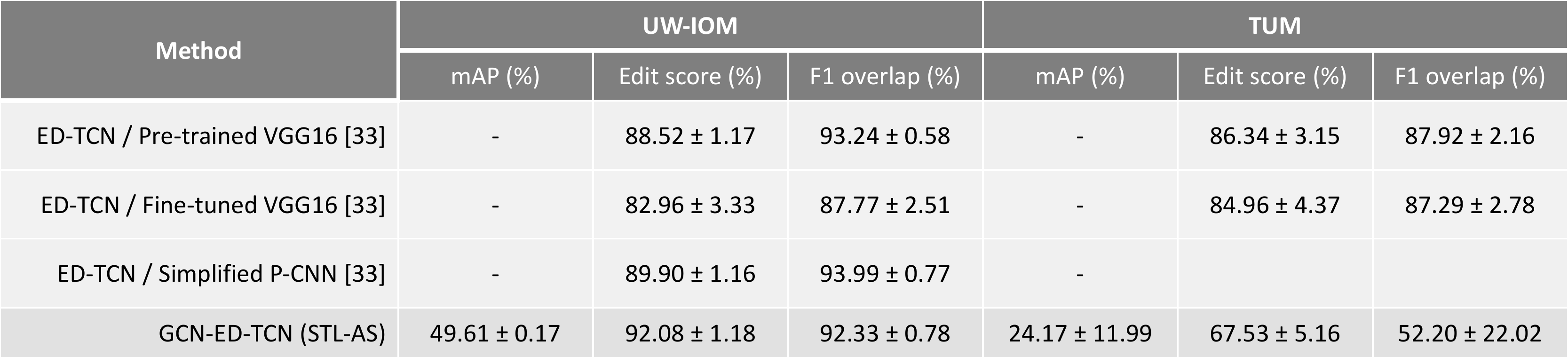}
\caption{mAP, edit, and F1-overlap score represented using mean and standard deviation values over the test videos in the UW-IOM and TUM datasets for different methods and modalities solving the HAS task.}
\label{tab:STL-Class}
\end{table*}
%-----------------------------------------------------------------------%
%
\begin{table}[H]
    \centering
    \includegraphics[width=1\columnwidth]{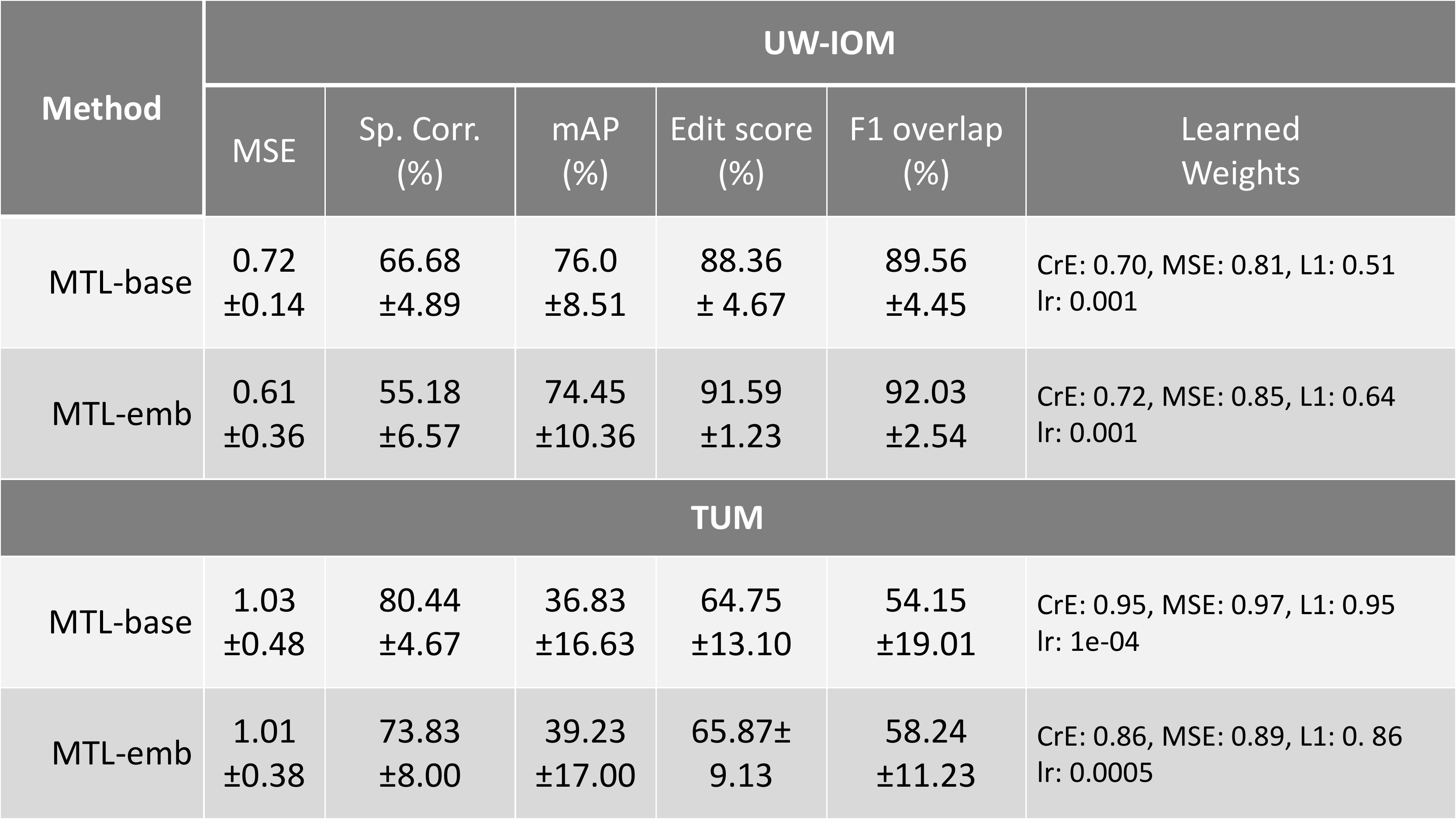}
    \caption{Results for the MTL network. mAP, edit, and F1-overlap scores are represented using mean and standard deviation values over the validation splits in the UW-IOM and TUM datasets for different activity segmentation methods and modalities. MSE and Spearman's coefficient show the model's performance in predicting the activity risk scores.}
    \label{tab:MTL}
\end{table}
\subsection{Fusion vs. No Fusion Approach}
\begin{figure}
\centering
  \includegraphics[width=.84\linewidth]{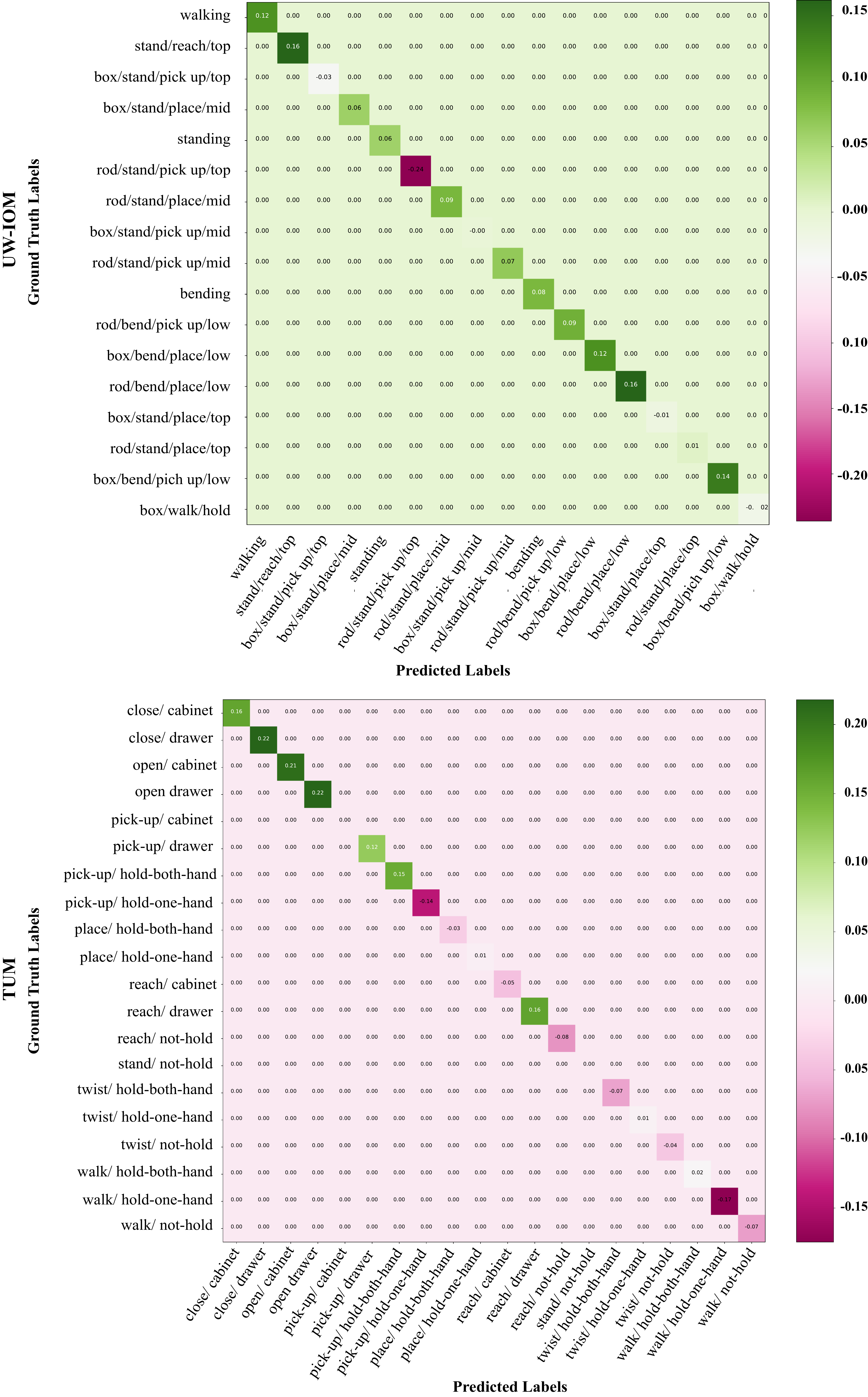}%
  \caption{The difference in confusion matrices. The top and bottom matrices are for the UW-IOM and TUM dataset, respectively. The diagonal elements show the differences between the diagonal values of the MTL-emb and MTL-base confusion matrices and the off-diagonal elements are shown as "0.0" for simplicity.}
  \label{fig:conf_diff}
\end{figure}

\noindent
The main purpose of this experiment is to validate the idea that action information can improve REBA score predictions. Table \ref{tab:MTL} and Figure \ref{fig:results} show the MTL-base and MTL-emb results, where improvements are observed when the \ac{HPA} head has access to the \texttt{Softmax} output of the \ac{HAS} head. In Figure \ref{fig:results}, we see the highly nonlinear ground truth REBA scoreline (in solid light blue-green) and the corresponding predictions for each action by both the \ac{MTL} networks. The figure suggests that the network with embedding predicts the REBA scores more accurately. On the contrary, the shared embedding model does not significantly improve the performance of the \ac{HAS} head. Figure \ref{fig:conf_diff} depicts the difference in the confusion matrices of the two models. For simplicity, the off-diagonal elements are ignored. While there are small  improvements in a few classes, the overall improvement is not substantial.
%////////////////////////////////////////////////////////////////////////
\subsection{Failure Cases}
\begin{figure}
\centering
  \includegraphics[width=.84\linewidth]{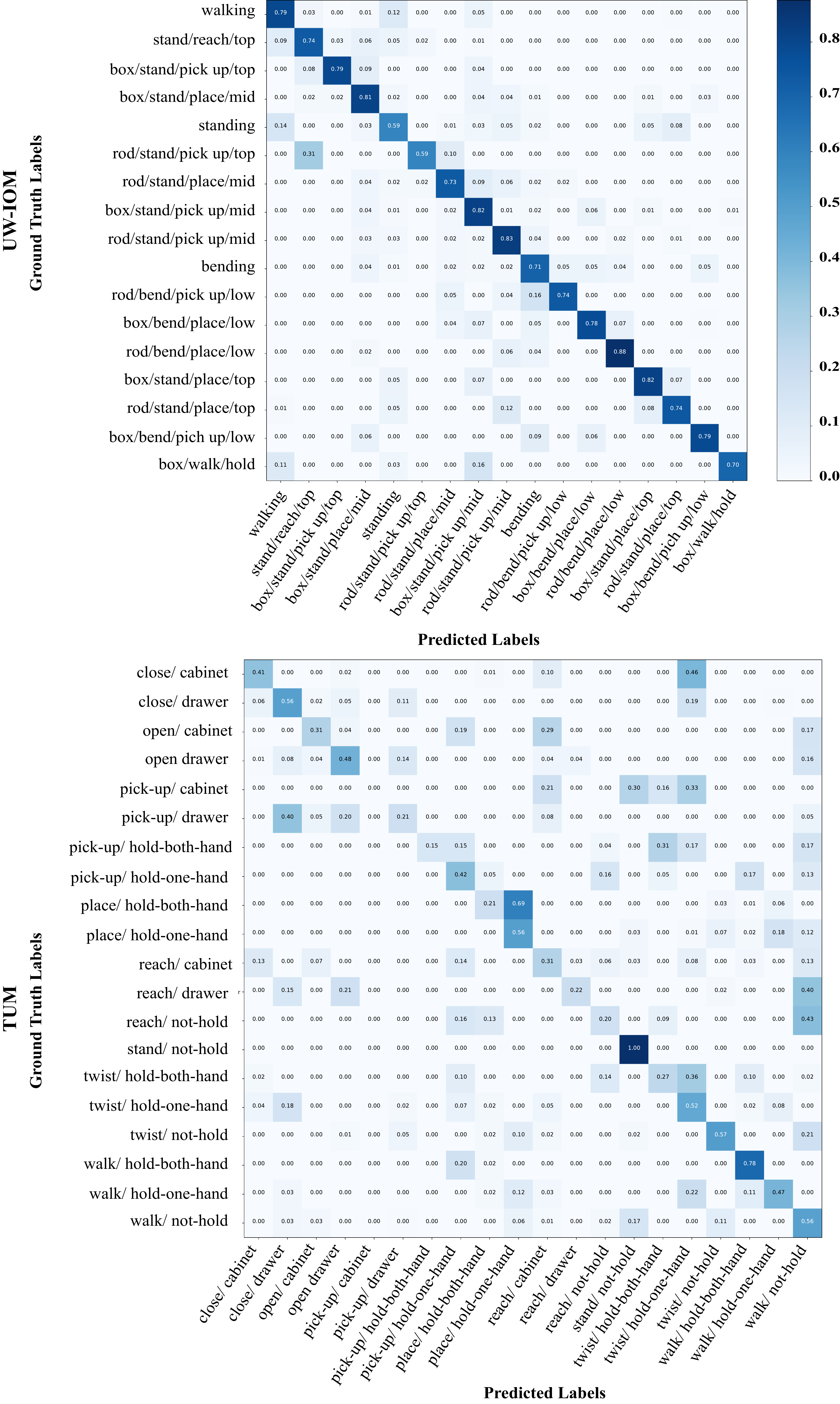}%
  \caption{Confusion matrices using MTL-base. The top and bottom matrices are for the UW-IOM and TUM dataset, respectively.}
  \label{fig:confs}
\end{figure}

\noindent
Although we show that our MTL-emb and STL-AS methods perform well on the UW-IOM dataset and even better than using context heavy features such as VGG16, these models are not particularly successful on the TUM dataset. We present the confusion matrices for the UW-IOM and TUM datasets in Figure \ref{fig:confs}. In the following, we describe our insights on the performance of the models in detail.

The camera view in the TUM dataset is from the top. As a result, arm pose estimation quality is poor for the activities where the person's back is facing the camera and the arm is occluded such as for \textit{pickup-drawer} and \textit{close-drawer}. Another source of confusion is between \textit{Pickup-hold-\textbf{both}-hands} and \textit{Pickup-hold-\textbf{one}-hands} due to the fact that the poses are very similar.

Since the segmentation head is not very successful on the TUM dataset, the improvement in the REBA score prediction between the MTL-emb and MTL-base models is also not significant unlike in the case of the UW-IOM dataset. For the TUM dataset, fusing image-based features with the GCN can be potentially useful in decreasing the ambiguity in the GCN spatial descriptors, thereby, improving both the STL-AS and MTL results for REBA score prediction.
%-----------------------------------------------------------------------%
%-----------------------------------------------------------------------%
%-----------------------------------------------------------------------%
%////////////////////////////////////////////////////////////////////////
\section{Conclusions and Future Work}
\noindent
We introduce a graph-based multi-task learning approach for \acl{HPA} and show that it outperforms the equivalent \acl{STL} due to the importance of the activity type in the risk associated with a posture. \acl{HPA} tasks, specifically \acl{ERA}, are more challenging than regular \acl{HAE} problems since the assessment has to happen in a frame-wise manner and is highly dependent on joint kinematics. Despite the challenge of tracking the intricacies of our risk assessment (REBA) profile, the proposed method shows competence in predicting the risk scores. More importantly, our work demonstrates the effectiveness of the \ac{GCN} model as a spatial feature extraction backbone, compared to context-based features that have been traditionally used with ED-TCN for \acl{HAS} tasks. To showcase the weaknesses of this framework, we implemented our method on a challenging dataset (TUM) and discussed the failure cases.

Although the focus of this work is on \acl{ERA}, we believe that our \acl{MTL} approach can be applied to many other action and skill assessment problems. The mapping of skeletal representation to the activity score using \ac{GCN} is a new approach for solving ERA, which can initiate a new direction by
%initiate a new path in many other activity assessment problems since there is a 
exploiting the natural connection between posture and activity risk/quality.

Although we outperform state-of-the-art on \ac{HAS} and \ac{ERA} on the UW-IOM dataset, some open issues remain. First, generalization concerning other activities has not been addressed. Our method learns the ergonomics risk scores in a supervised learning framework, which makes the performance of the model limited to the labeled activities that have been observed. Second, only joint positions are considered in the spatial representation, while other kinematic information such as velocity and acceleration have been shown to be important for many types of injury such as back injuries \cite{marras2009loading}. In the future, we hope to address these issues by developing a biomechanics-based human pose representation model that learns the causal relation between joint kinematics and the resultant ergonomics risk.
{\small
\bibliographystyle{ieee_fullname}
\bibliography{References}
}
\end{document}